\documentclass[10pt,conference,letterpaper]{IEEEtran}
\usepackage{times,amsmath,epsfig}
\usepackage{epsfig}
\usepackage{latexsym}
\usepackage{amsfonts}
\usepackage{xspace}
\usepackage{amssymb}
\usepackage{graphicx}
\usepackage{psfrag}
\usepackage{subfigure}
\usepackage{url}
\usepackage{amsmath}
\usepackage{array}
\usepackage{pstricks}
\usepackage{algorithm}
\usepackage{algorithmic}
\usepackage{graphicx}
\DeclareGraphicsExtensions{.pdf,.jpg,.png,.tif}
\title{A Review of Verbal and Non-Verbal Human-Robot Interactive Communication}
\author{%
{Nikolaos Mavridis}%
\vspace{1.6mm}\\
\fontsize{10}{10}\selectfont\itshape
Interactive Robots and Media Lab, NCSR Demokritos\\
GR-15310, Agia Paraskevi, Athens, Greece\\                                                                                                                                                                                                                                                                                                                                                                                                                                                                                                                                                                                                                                                                                                                                                                                                                                                                                                                                                                                                                                                                                                                                                                                                                                                                                                                                                            
\vspace{1.2mm}\\
\fontsize{9}{9}\selectfont\ttfamily\upshape
$\{$nmav$\}$@alum.mit.edu\\
}
\begin{document}
\maketitle
\begin{abstract}
In this paper, an overview of human-robot interactive communication is presented, covering verbal as well as non-verbal aspects. Following a historical introduction, and motivation towards fluid human-robot communication, ten desiderata are proposed, which provide an organizational axis both of recent as well as of future research on human-robot communication. Then, the ten desiderata are examined in detail, culminating to a unifying discussion, and a forward-looking conclusion.
\end{abstract}

\section{Introduction: Historical Overview}\label{intro}
While the first modern-day industrial robot, Unimate, began work on the General Motors assembly line in 1961, and was conceived in 1954 by George Devol \cite{DevolObituary,devol1984encoding}, the concept of a robot has a very long history, starting in mythology and folklore, and the first mechanical predecessors (automata) having been constructed in Ancient Times. For example, in Greek mythology, the God Hephaestus is reputed to have made mechanical servants from gold (\cite{Gera} in p.114, and \cite{Iliad} verse 18.419). Furthermore, a rich tradition of designing and building mechanical, pneumatic or hydraulic automata also exists: from the automata of Ancient Egyptian temples, to the mechanical pigeon of the Pythagorean Archytas of Tarantum circa 400BC \cite{huffman2005archytas}, to the accounts of earlier automata found in the Lie Zi text in China in 300BC \cite{needham1959}, to the devices of Heron of Alexandria \cite{Sharkey} in the 1st century. The Islamic world also plays an important role in the development of automata; Al-Jazari, an Arab inventor, designed and constructed numerous automatic machines, and is even reputed to have devised the first programmable humanoid robot in 1206AD \cite{Rosheim}. The word ``robot'', a Slavic word meaning servitude, was first used in this context by the Czech author Karel Capek in 1921 \cite{Capek}.

However, regarding robots with natural-language conversational abilities, it wasn’t until the 1990's that the first pioneering systems started to appear. Despite the long history of mythology and automata, and the fact that even the mythological handmaidens of Hephaestus were reputed to have been given a voice \cite{Gera}, and despite the fact that the first general-purpose electronic speech synthesizer was developed by Noriko Omeda in Japan in 1968 \cite{Klatt}, it wasn’t until the early 1990's that conversational robots such as MAIA \cite{Antoniol}, RHINO \cite{Rhino}, and AESOP \cite{Versweyveld} appeared. These robots cover a range of intended application domains; for example, MAIA was intended to carry objects and deliver them, while RHINO is a museum guide robot, and AESOP a surgical robot. 

In more detail, the early systems include Polly, a robotic guide that could give tours in offices \cite{Horswill93polly, Polly1996}. Polly had very simple interaction capacities; it could perceive human feet waving a ``tour wanted'' signal, and then it would just use pre-determined phrases during the tour itself. A slightly more advanced system was TJ \cite{Torrance}. TJ could verbally respond to simple commands, such as ``go left'', albeit through a keyboard. RHINO, on the other hand \cite{Rhino}, could respond to tour-start commands, but then, again, just offered a pre-programmed tour with fixed programmer-defined verbal descriptions. Regarding mobile assistant robots with conversational capabilities in the 1990s, a classic system is MAIA \cite{Antoniol,Antoniol94}, obeying simple commands, and carrying objects around places, as well as the mobile office assistant which could not only deliver parcels but guide visitors described in \cite{Ant1996}, and the similar in functionality Japanese-language robot Jijo-2 \cite{Asoh99,FAM1998,Matsui:1999}. Finally, an important book from the period is \cite{crangle1994language}, which is characteristic of the traditional natural-language semantics-inspired theoretical approaches to the problem of human-robot communication, and also of the great gap between the theoretical proposals and the actual implemented systems of this early decade. 

What is common to all the above early systems is that they share a number of limitations. First, all of them only accept a fixed and small number of simple \textit{canned commands}, and they respond with a set of \textit{canned answers}. Second, the only \textit{speech acts} (in the sense of Searle~\cite{Searle:1969:SPA}) that they can handle are requests. Third, the dialogue they support is clearly not flexibly \textit{mixed initiative}; in most cases it is just human-initiative. Four, they don’t really support \textit{situated language}, i.e. language about their physical situations and events that are happening around them; except for a fixed number of canned location names in a few cases. Five, they are not able to handle \textit{affective speech}; i.e. emotion-carrying prosody is neither recognized nor generated. Six, their \textit{non-verbal communication} \cite{vogeley2010artificial} capabilities are almost non-existent; for example, gestures, gait, facial expressions, and head nods are neither recognized nor produced. And seventh, their dialogue systems are usually effectively stimulus-response or stimulus-state-response systems; i.e. no real \textit{speech planning} or purposeful dialogue generation is taking place, and certainly not in conjunction with the motor planning subsystems of the robot. Last but quite importantly, no real \textit{learning}, off-line or on-the-fly is taking place in these systems; verbal behaviors have to be prescribed. 

All of these shortcomings of the early systems of the 1990s, effectively have become desiderata for the next two decades of research: the 2000s and 2010s, which we are in at the moment. Thus, in this paper, we will start by providing a discussion giving motivation to the need for existence of interactive robots with natural human-robot communication capabilities, and then we will enlist a number of desiderata for such systems, which have also effectively become areas of active research in the last decade. Then, we will examine these desiderata one by one, and discuss the research that has taken place towards their fulfillment. Special consideration will be given to the so-called ``symbol grounding problem'' \cite{harnad1990symbol}, which is central to most endeavors towards natural language communication with physically embodied agents, such as robots. Finally, after a discussion of the most important open problems for the future, we will provide a concise conclusion.

\section{Motivation: Interactive Robots with Natural Language capabilities – but why?}\label{sec:Motivation}

There are at least two avenues towards answering this fundamental question, and both will be attempted here. The first avenue will attempt to start from first principles – and derive a rationale towards equipping robots with natural language. The second, more traditional and safe avenue, will start from a concrete, yet partially transient, base: application domains – existing or potential. In more detail:

Traditionally, there used to be clear separation between design and deployment phases for robots. Application-specific robots (for example, manufacturing robots, such as \cite{Kuka2010}) were: (a) designed by expert designers, (b) possibly tailor-programmed and occasionally reprogrammed by specialist engineers at their installation site, and (c) interacted with their environment as well as with specialized operators during actual operation. However, the phenomenal simplicity but also the accompanying inflexibility and cost of this traditional setting is often changing nowadays. For example, one might want to have broader-domain and less application-specific robots, necessitating more generic designs, as well as less effort by the programmer-engineers on site, in order to cover the various contexts of operation. Even better, one might want to rely less on specialized operators, and to have robots interact and collaborate with non-expert humans with little if any prior training. Ideally, even the actual traditional programming and re-programming might also be transferred over to non-expert humans; and instead of programming in a technical language, to be replaced by intuitive tuition by demonstration, imitation and explanation \cite{wrede2013user, argall2009survey, nehaniv2007imitation}. Learning by demonstration and imitation for robots already has quite some active research; but most examples only cover motor and aspects of learning, and language and communication is not involved deeply.

And this is exactly where natural language and other forms of fluid and natural human-robot communication enter the picture: Unspecialized non-expert humans are used to (and quite good at) teaching and interacting with other humans through a mixture of natural language as well as nonverbal signs. Thus, it makes sense to capitalize on this existing ability of non-expert humans by building robots that do not require humans to adapt to them in a special way, and which can fluidly collaborate with other humans, interacting with them and being taught by them in a natural manner, almost as if they were other humans themselves. 

Thus, based on the above observations, the following is one classic line of motivation towards justifying efforts for equipping robots with natural language capabilities: Why not build robots that can comprehend and generate human-like interactive behaviors, so that they can cooperate with and be taught by non-expert humans, so that they can be applied in a wide range of contexts with ease? And of course, as natural language plays a very important role within these behaviors, why not build robots that can fluidly converse with humans in natural language, also supporting crucial non-verbal communication aspects, in order to maximize communication effectiveness, and enable their quick and effective application?

Thus, having presented the classical line of reasoning arriving towards the utility of equipping robots with natural language capabilities, and having discussed a space of possibilities regarding role assignment between human and robot, let us now move to the second, more concrete, albeit less general avenue towards justifying conversational robots: namely, specific applications, existing or potential. Such applications, where natural human-robot interaction capabilities with verbal and non-verbal aspects would be desirable, include: flexible manufacturing robots; lab or household robotic assistants \cite{mavridis2006grounded, van2007robocup, Foster2006GermanRobots, giuliani2011evaluating}; assistive robotics and companions for special groups of people \cite{wada2007paro}; persuasive robotics (for example, \cite{kamei2010recommendation}); robotic receptionists \cite{roboreceptionistarab}, robotic educational assistants, robotic wheelchairs \cite{tellex2006spatial}, companion robots \cite{dautenhahn2006may}, all the way to more exotic domains, such as robotic theatre actors \cite{ mavridis2009ibnsina}, musicians \cite{petersen2010musical}, dancers \cite{kosuge2003dance} etc. 

In all of the above applications, although there is quite some variation regarding requirements, one aspect at least is shared: the desirability of natural fluid interaction with humans supporting natural language and non-verbal communication, possibly augmented with other means. Of course, although this might be desired, it is not always justified as the optimum choice, given technico-economic constraints of every specific application setting. A thorough analysis of such constraints together with a set of guidelines for deciding when natural-language interaction is justified, can be found at \cite{kulyukin2006natural}. 

Now, having examined justifications towards the need for natural language and other human-like communication capabilities in robots across two avenues, let us proceed and become more specific: natural language, indeed – but what capabilities do we actually need?

\section{Desiderata - What might one need from a conversational Robot?}\label{sec:desiderata}

An initial list of desiderata is presented below, which in neither totally exhuastive nor absolutely orthogonal; however, it serves as a good starting point for discussing the state of the art, as well as the potentials of each of the items:

D1) Breaking the ``simple commands only'' barrier

D2) Multiple speech acts

D3) Mixed initiative dialogue

D4) Situated language and the symbol grounding problem

D5) Affective interaction

D6) Motor correlates and Non-Verbal Communication

D7) Purposeful speech and planning

D8) Multi-level learning

D9) Utilization of online resources and services

D10) Miscellaneous abilities 

The particular order of the sequence of desiderata, was chosen for the purpose of illustration, as it provides partially for a building-up of key points, also allowing for some tangential deviations. 

\subsection{Breaking the ``simple commands only'' barrier}\label{sec:breaking}

The traditional conception of conversational robots, as well as most early systems, is based on a clear human-master robot-servant role assignment, and restricts the robots conversational competencies to simple ``motor command requests'' only in most cases. A classic example can be seen for example in systems such as \cite{mavridis2006grounded}, where a typical dialogue might be:

H: ``Give me the red one''

R: (Picks up the red ball, and gives to human)

H: ``Give me the green one''

R: ``Do you mean this one, or that one?'' (robot points to two possible candidate objects)

H: ``The one on the left''

R: (Picks up the green ball on the left, and hands over to human)

What are the main points noticing in this example? Well, first of all, (p1) this is primarily a single-initiative dialogue: the human drives the conversation, the robot effectively just producing motor and verbal responses to the human verbal stimulus. Second, (p2) apart from some disambiguating questions accompanied by deixis, there is not much that the robot says – the robot primarily responds with motor actions to the human requests, and does not speak. And, (p3) regarding the human statements, we only have one type of speech acts \cite{Searle:1969:SPA}: RequestForMotorAction. Furthermore, (p4) usually such systems are quite inflexible regarding multiple surface realizations of the acceptable commands; i.e. the human is allowed to say ``Give me the red one'', but if he instead used the  elliptical ``the red object, please'' he might have been misinterpreted and (p5) in most cases, the mapping of words-to-responses is arbitrarily chosen by the designer; i.e. motor verbs translate to what the designer thinks they should mean for the robot (normative meaning), instead of what an empirical investigation would show regarding what other humans would expect they mean (empirical meaning).

Historically, advanced theorization for such systems exists as early as \cite{crangle1994language}, and there is still quite a stream of active research which, although based on beautiful and systematic formalizations and eloquent grammars, basically produces systems which would still fall within the three points mentioned above. Such an example is \cite{dzifcaketal09icra}, in which a mobile robot in a multi-room environment, can handle commands such as: ``Go to the breakroom and report the location of the blue box''

Notice that here we are not claiming that there is no importance in this research that falls within this strand; we are just mentioning that, as we shall see, there are many other aspects of natural language and robots, which are left unaccounted by such systems. Furthermore, it remains to be seen, how many of these aspects can later be effectively integrated with systems belonging to this strand of research.

\subsection{Multiple speech acts}\label{sec:speechacts}

The limitations (p1)-(p5) cited above for the classic ``simple commands only'' systems provide useful departure points for extensions. Speech act theory was introduced by J.L.Austin \cite{austin1962tw}, and a speech act is usually defined as an utterance that has performative function in language and communication. Thus, we are focusing on the function and purpose of the utterance, instead of the content and form. Several taxonomies of utterances can be derived according to such a viewpoint: for example, Searle \cite{Searle75}, proposed a classification of illocutionary speech acts into assertives, directives, commisives, expressives, and declarations. Computational models of speech acts have been proposed for use in human-computer interaction \cite{Allen01towardsconversational}.

In this light of speech acts, lets us start by extending upon point (p3) made in the previous section. In the short human-robot dialogue presented in the previous section, the human utterances ``Give me the red one'' and ``Give me the green one'' could be classified as Request speech acts, and more specifically requests for motor action (one could also have requests for information, such as ``What color is the object?'' etc.). But what else might one desire in terms of speech act handling capabilities, apart from RequestForMotorAction (which we shall call SA1, a Directive according to \cite{Searle75})? Some possibilities follow below:

H: ``How big is the green one?'' (RequestForInformAct, SA2, Directive)

H: ``There is a red object at the left'' (Inform, SA3, Assertive)

H: ``Let us call the small doll Daisy'' (Declare, SA4, Declaration)

And many more exist. Systems such as \cite{mavridis2007grounded} are able to handle SA2 and SA3 apart from SA1-type acts; and one should also notice, that there are many classificatory systems for speech acts, across different axis of classification, and with multiple granularities. Also, it is worth starting at this stage to contemplate upon what might it mean to respond appropriately to different kinds of speech acts. For example, an appropriate response to a RequestForMotorAction (a Directive) is the motor action itself, if unambiguous and feasible; however, an appropriate response to an Assertive or a Declarative consists of a change to some form of a ``mental model'' \cite{Johnson1983mental} or ``situation model'' \cite{zwaan1998situation} \cite{mavridis2007grounded} that the robot might be keeping; i.e. creating an appropriate mental token for an object in the case of ``There is a red object at the left'', or changing the name label for a mental object token in the case of ``Let us call this small doll Daisy''; i.e. both statements elicit primarily internal (mental) actions, instead of external (motor or verbal) actions. 

Another relevant aspect of speech act theory is the handling of indirect speech acts.  For example, consider the following utterance:

H: ``Ah, it is quite hot in this room'' (phenomenally, an Assertive),

Which might actually be a polite way of saying: 

H: ``Open the window'' (essentially, a Directive).

This substitution of an Assertive for an implied Directive (to be inferred by the listener) is a classic example of an indirect speech act. Usually, the analysis of such acts is based on the Gricean maxims of conversation \cite{grice1975logic}; and numerous computational implementations for handling such indirect speech acts have been proposed, such as \cite{wilske2006service}.

Finally, yet another problem related to speech acts, is the issue of their classification from the robot, after hearing them. Classic techniques such as those described on \cite{krsmanovic2006have} rely on liguistic information only; however, paralinguistic information (such as prosodic features) can also prove useful towards speech act classification; the interested reader is referred for example to \cite{ishi2006analysis}.

\subsection{Mixed Initiative Dialogue}\label{sec:initiative}

Now, starting again from the shortcoming of the traditional ``simple commands-only'' systems, let us extend across another axis, namely (p1): human-initiative dialogue is not the only option; one could also have robot-initiative, or ideally, full mixed-initiative. Consider FaceBots \cite{mavridis2010facebots} \cite{mavridis2009facebots}, a conversational robot utilizing facebook-derived information. A typical dialogue might include turns such as:

R: ``Hello! Are you Noura Dhaheri?''

H: ``Yes''

R: ``Hello Noura! Are you doing well today?''

H: ``Pretty well!''

R: ``Great! Do you know that our friend Wajahat Kazmi has posted a new photo?''

H: ``No!''

R: ``Well, I am sending you an inbox message for this, check it out!''

This is a classic example of robot-initiative dialogue, and actually with very simple responses expected from the human – mainly Yes or No. On the other hand, consider a dialogue such as (from the robot BIRON at Bielefeld, dialogue in \cite{wrede2006analyses}), which is an example of a dialogue with limited mixed initiative:

R: ``Hello Human!''

H: ``Hello! Who are you?''

R: ``My name is BIRON, if you show me something I can remember it.''

H: ``OK, um, look, this is a mug.''

R: ``I beg your pardon?''

H: ``This is a mug, umm, here.''

R: ``Sorry, can you use some other words?''

H: ``This is a mug!''

R: ``Sorry, I cannot find it. Are you pointing to it with your finger?''

H: ``Yes!''

R: ``I know it is sometimes difficult with me, but don’t feel discouraged!''

H: (laugh)

R: ``OK, I have found it, it is really nice!''

Here, it is neither the robot nor the human driving the dialogue all of the time; for example, the opening pair, R-H is robot-initiative (R: ``Hello Human!'' H: ``Hello!''). However, directly afterwards, the second half of H together with R create a second pair of exchanges, which is human-initiative (H: ``Who are you?'', R: ``My names is BIRON...''). And thus the initiative can be reversed in multiple points throughout the dialogue.

For an investigation of the state of the art towards mixed initiative, the interested reader is referred to examples such as the Karlsruhe Humanoid \cite{stiefelhagen2007enabling}the Biron and Barthoc systems at Bielefeld \cite{wrede2006analyses}, and also workshops such as \cite{ertlimproving}. 

\subsection{Situated Language and Symbol Grounding}\label{sec:situatedsymbol}

Yet another observation regarding shortcomings of the traditional command-only systems that is worth extending from, was point (p5) that was mentioned above: the meanings of the utterances were normatively decided by the designer, and not based on empirical observations. For example, a designer/coder could normatively pre-define the semantics of the color descriptor ``red'' as belonging to the range between two specific given values. Alternatively, one could empirically get a model of the applicability of the descriptor ``red'' based on actual human usage; by observing the human usage of the word in conjunction with the actual apparent color wavelength and the context of the situation. Furthermore, the actual vocabularies (“red”, ``pink'', etc.) or the classes of multiple surface realizations (p4) (quasi-synonyms or semantically equivalent parts of utterances, for example: ``give me the red object'', ``hand me the red ball''), are usually hand-crafted in such systems, and again not based on systematic human observation or experiment. 

There are a number of notable exceptions to this rule, and there is a growing tendancy to indeed overcome these two limitations recently. For example, consider \cite{ralph2008toward}, during which a wizard-of-oz experiment provided the collection of vocabulary from users desiring to verbally interact with a robotic arm, and examples such as \cite{tellex2006spatial}, for which the actual context-depending action models corresponding to simple verbal commands like ``go left'' or ``go right'' (which might have quite different expected actions, depending on the surrounding environment) were learnt empirically through human experiments. 

Embarking upon this avenue of thought, it slowly becomes apparent that the connection between local environment (and more generally, situational context) and procedural semantics of an utterance is quite crucial. Thus, when dealing with robots and language, it is impossible to isolate the linguistic subsystems from perception and action, and just plug-and-play with a simple speech-in speech-out black box chatterbot of some sort (such as the celebrated ELIZA \cite{ weizenbaum1966eliza} or even the more recent victors of the Loebner Prize \cite{ mauldin1994chatterbots}). Simply put, in such systems, there is no connection of what is being heard or said to what the robot senses and what the robot does. This is quite a crucial point; there is a fundamental need for closer integration of language with sensing, action, and purpose in conversational robots \cite{mavridis2006grounded} \cite{mavridis2007grounded}, as we shall also see in the next sections. 

\subsubsection{Situated Language}\label{sec:situated}

Upon discussing the connection of language to the physical context, another important concept becomes relevant: situated language, and especially the language that children primarily use during their early years; i.e. language that is not abstract or about past or imagined events; but rather concrete, and about the physical here-and-now. But what is the relevance of this observation to conversational robots? One possibility is the following; given that there seems to be a progression of increasing complexity regarding human linguistic development, often in parallel to a progression of cognitive abilities, it seems reasonable to: First partially mimic the human developmental pathway, and thus start by building robots that can handle such situated language, before moving on to a wider spectrum of linguistic abilities. This is for example the approach taken at  \cite{mavridis2007grounded}.

Choosing situated language as a starting point also creates a suitable entry point for discussing language grounding in the next section. Now, another question that naturally follows is: could one postulate a number of levels of extensions from language about the concrete here-and-now to wider domains? This is attempted in \cite{mavridis2007grounded}, and the levels of increasing detachment from the ``here-and-now'' postulated there are:

First level: limited only to the ``here-and-now, existing concrete things''. Words connect to things directly accessible to the senses at the present moment. If there is a chair behind me, although I might have seen it before, I cannot talk about it - ``out of sight'' means ``non-existing'' in this case. For example, such a robotic system is \cite{roy2000computational}

Second level: (``now, existing concrete things''); we can talk about the ``now'', but we are not necessarily limited to the ``here'' - where here means currently accessible to the senses. We can talk about things that have come to our senses previously, that we conjecture still exist through some form of psychological ``object permanence'' \cite{baillargeon1985object} - i.e., we are keeping some primitive ``mental map'' of the environment. For example, this was the state of the robot Ripley during \cite{roy2004mental}

Third level: (``past or present, existing concrete things''), we are also dropping the requirement of the ``now'' - in this case, we also posses some form of episodic memory \cite{tulving1983elements} enabling us to talk about past states. An example robot implementation can be found in \cite{mavridis2010human}

Fourth level: (``imagined or predicted concrete things''); we are dropping the requirement of actual past or present existence, and we can talk about things with the possibility of actual existence - either predicted (connectible to the present) or imagined. \cite{mavridis2007grounded}

Fifth level: (``abstract things'') we are not talking about potentially existing concrete things any more, but about entities that are abstract. But what is the criterion of ``concreteness''? A rough possibility is the following: a concrete thing is a first-order entity (one that is directly connected to the senses); an ``abstract'' thing is built upon first order entities, and does not connect directly to the senses, as it deals with relationships between them. Take, for example, the concept of the ``number three'': it can be found in an auditory example (``threeness'' in the sound of three consecutive ticks); it can also be found in a visual example (``threeness'' in the snapshot of three birds sitting on a wire). Thus, threeness seems to be an abstract thing (not directly connected to the senses).

Currently, there exist robots and methodologies \cite{mavridis2007grounded} that can create systems handling basic language corresponding to the first four stages of detachment from situatedness; however, the fifth seems to still be out of reach. If what we are aiming towards is a robot with a deeper understanding of the meaning of words referring to abstract concepts, although related work on computational analogy making (such as \cite{gentner2011computational}), could prove to provide some starting points for extensions towards such domains, we are still beyond the current state-of-the-art.

Nevertheless, there are two interesting points that have arisen in the previous sections: first, that when discussing natural language and robots, there is a need to connect language not only to sensory data, but also to internalized ``mental models'' of the world – in order for example to deal with detachment from the immediate ``here-and-now''. And second, that one needs to consider not only phonological and syntactical levels of language – but also questions of semantics and meaning; and pose the question: ``what does it mean for a robot to understand a word that it hears or utters''? And also, more practically: what are viable computational models of the meaning of words, suitable to embodied conversational robots? We will try to tackle these questions right now, in the next subsection.

\subsubsection{Symbol Grounding}\label{sec:symbol}

One of the main philosophical problems that arises when trying to create embodied conversational robots is the so-called ``symbol grounding problem'' \cite{harnad1990symbol}. In simple terms, the problem is the following: imagine a robot, having an apple in front of it, and hearing the word ``apple'' –  a verbal label which is a conventional sign (in semiotic terms \cite{pierce1955logic} \cite{peirce1974collected}), and which is represented by a symbol within the robots cognitive system. Now this sign is not irrelevant to the actual physical situation; the human that uttered the word ``apple'' was using it to refer to the physical apple that is in front of the robot. Now the problem that arises is the following: how can we connect the symbol standing for ``apple'' in the robots cognitive system, with the physical apple that it refers to? Or, in other words, how can we ground out the meaning of the symbol to the world? In simple terms, this is an example of the symbol grounding problem. Of course, it extends not only to objects signified by nouns, but to properties, relations, events etc., and there are many other extensions and variations of it.

So, what are solutions relevant to the problem? In the case of embodied robots, the connection between the internal cognitive system of the robot (where the sign is) and the external world (where the referent is) is mediated through the sensory system, for this simple case described above. Thus, in order to ground out the meaning, one needs to connect the symbol to the sensory data – say, to vision. Which is at least, to find a mechanism through which, achieves the following bidirectional connection: first, when an apple appears in the visual stream, instantiates an apple symbol in the cognitive system (which can later for example trigger the production of the word ``apple'' by the robot), and second, when an apple symbol is instantiated in the cognitive system (for example, because the robot heard that ``there is an apple''), creates an expectation regarding the contents of the sensory stream given that an apple is reported to be present. This bidirectional connection can be succinctly summarized as:

\scriptsize{external referent $>$ sensory stream $>$ internal symbol $>$ produced utterance}

external referent $<$ sensory expectation $<$ internal symbol $<$ heard utterance

\normalsize
This bidirectional connection we will refer to as ``full grounding'', while its first unidirectional part as ``half grounding''. Some notable papers presenting computational solutions of the symbol grounding problem for the case of robots are: half-grounding of color and shapes for the Toco robot \cite{roy2000computational}, and full-grounding of multiple properties for the Ripley robot \cite{mavridis2006grounded}. Highly relevant work includes: \cite{spexard2006biron} and also Steels \cite{steels2003evolving}, \cite{larson2004intrinsic}, \cite{gorniak2005affordance}, and also \cite{ yu2008grounding} from a child lexical perspective.

The case of grounding of spatial relations (such as ``to the left of'', ``inside'' etc.) reserves special attention, as it is a significant field on its own. A classic paper is \cite{regier2001grounding}, presenting an empirical study modeling the effect of central and proximal distance on 2D spatial relations; regarding the generation and interpretation of referring expressions on the basis of landmarks for a simple rectangle world, there is \cite{roy2002learning}, while the book by \cite{ coventry2004saying} extends well into illustrating the inadequacy of geometrical models and the need for functional models when grounding terms such as ``inside'', and covers a range of relevant interesting subjects. Furthermore, regarding the grounding of attachment and support relations in videos, there is the classic work by \cite{siskind2011grounding}. For an overview of recent spatial semantics research, the interested reader is referred to \cite{zlatev2007spatial}, and a sampler of important current work in robotics includes \cite{skubic2004spatial}, \cite{zender2008conceptual}, \cite{tellex2011understanding}, and the most recent work of Tellex on grounding with probabilistic graphical models \cite{tellex2011approaching}, and for learning word meanings from unaligned parallel data \cite{tellex2013learning}.

Finally, an interesting question arises when trying to ground out personal pronouns, such as ``me, my, you, your''. Regarding their use as modifiers of spatial terms (``my left''), relevant work on a real robot is \cite{roy2004mental}, and regarding more general models of their meaning, the reader is referred to \cite{gold2006grounded}, where a system learns the semantics of the pronouns through examples.

A number of papers has recently also appeared claiming to have provided a solution to the ``symbol grounding problem'', such as \cite{steels2008symbol}. There is a variety of different opinions regarding what an adequate solution should accomplish, though. A stream of work around an approach dealing with the evolution of language and semiotics, is outlined in \cite{steels2006semiotic}. From a more applied and practical point of view though, one would like to be able to have grounded ontologies \cite{hudelot2005symbol} \cite{cregan2007symbol} or even robot-usable lexica augmented with computational models providing such grounding: and this is the ultimate goal of the EU projects POETICON \cite{wallraven2011poeticon} \cite{pastra2010poeticon}, and the follow-up project POETICON II.

Another important aspect regarding grounding is the set of qualitatively different possible target meaning spaces for a concept. For example, \cite{mavridis2007grounded} proposes three different types of meaning spaces: sensory, sensorymotor, and teleological. A number of other proposals exists for meaning spaces in cognitive science, but not directly related to grounding; for example, the geometrical spaces Gardenfors \cite{gardenfors2004conceptual}. Furthermore, any long-ranging agenda towards extending symbol grounding to an ever-increasing range of concepts, needs to address yet another important point: semantic composition, i.e. for a very simple example, consider how a robot could combine a model of ``red'' with a model of ``dark'' in order to derive a model of ``dark red''. Although this is a fundamental issue, as discussed in \cite{mavridis2007grounded}, it has yet to be addressed properly.

Last but not least, regarding the real-world acquisition of large-scale models of grounding in practice, special data-driven models are required, and the quantities of empirical data required would make collection of such data from non-experts (ideally online) highly desirable. Towards that direction, there exists the pioneering work of Gorniak \cite{gorniak2005affordance} where a specially modified computer game allowed the collection of referential and functional models of meaning of the utterances used by the human players. This was followed up by \cite{orkin2007restaurant} \cite{chernova2010crowdsourcing} \cite{depalma2011leveraging}, in which specially designed online games allowed the acquisition of scripts for situationally appropriate dialogue production. These experiments can be seen as a special form of crowdsourcing, building upon the ideas started by pioneering systems such as Luis Von Ahn’s peekaboom game \cite{von2006peekaboom}, but especially targeting the situated dialogic capabilities of embodied agents. Much more remains to be done in this promising direction in the future.

\subsubsection{Meaning Negotiation}\label{sec:negotiation}

Having introduced the concept of non-logic-like grounded models of meaning, another interesting complication arises. Given that different conversational partners might have different models of meaning, say for the lexical semantics of a color term such as ``pink'', how is communication possible? A short, yet minimally informative answer, would be: given enough overlap of the particular models, there should be enough shared meaning for communication. But if one examines a number of typical cases of misalignment across models, he will soon reach to the realization that models of meaning, or even second-level models (beliefs about the models that others hold), are very often being negotiated and adjusted online, during a conversation. For example:

(Turquoise object on robot table, in front of human and robot)

H: ``Give me the blue object!''

R: ``No such object exists''

H: ``Give me the blue one!''

R: ``No such object exists''

But why is this surreal human-robot dialog taking place, and why it would not have taken place for the case of two humans in a similar setting? Let us analyze the situation. The object on the table is turquoise, a color which some people might classify as ``blue'', and others as ``green''. The robot’s color classifier has learnt to treat turquoise as green; the human classifies the object as ``blue''. Thus, we have a categorical misalignment error, as defined in \cite{mavridis2007grounded}. For the case of two humans interacting instead of a human and a robot, given the non-existence of another unique referent satisfying the ``blue object'' description, the second human would have readily assumed that most probably the first human is classifying turquoise as ``blue''; and, thus, he would have temporarily adjusted his model of meaning for ``blue'' in order to be able to include turquoise as ``blue'', and thus to align his communication with his conversational partner. Thus, ideally we would like to have conversational robots that can gracefully recover from such situations, and fluidly negotiate their models of meaning online, in order to be able to account for such situations. Once again, this is a yet unexplored, yet crucial and highly promising avenue for future research.

\subsection{Affective Interaction}\label{sec:affective}

An important dimension of cognition is the affective/emotional. In the german psychological tradition of the 18th century, the affective was part of the tripartite classification of mental activities into cognition, affection, and conation; and apart from the widespread use of the term, the influence of the tri-partite division extended well into the 20th century \cite{hilgard1980trilogy}.

The affective dimension is very important in human interaction \cite{picard2003affective}, because it is strongly intertwined with learning \cite{picard2004affective}, persuasion \cite{haddock2008should}, and empathy, among many other functions. Thus, it carries over its high significance for the case of human-robot interaction. For the case of speech, affect is marked both in the semantic/pragmatic content as well as in the prosody of speech: and thus both of these ideally need to be covered for effective human-robot interaction, and also from both the generation as well as recognition perspectives. Furthermore, other affective markers include facial expressions, body posture and gait, as well as markers more directly linked to physiology, such as heart rate, breathing rate, and galvanic skin response. 

Pioneering work towards affective human-robot interaction includes \cite{breazeal2003emotion} where, extending upon analogous research from virtual avatars such as Rea \cite{cassell2000embodied}, Steve \cite{johnson2000animated}, and Greta \cite{rosis2003greta}, Cynthia Breazeal presents an interactive emotion and drive system for the Kismet robot \cite{ breazeal1998toward}, which is capable of multiple facial expressions. An interesting cross-linguistic emotional speech corpus arising from children’s interactions with the Sony AIBO robot is presented in \cite{batliner2004you}. Another example of preliminary work based on a Wizard-of-Oz approach, this time regarding children’s interactions with the ATR Robovie robot in Japan, is presented in \cite{komatani2004recognition}. In this paper, automatic recognition of embarrassment or pleasure of the children is demonstrated. Regarding interactive affective storytelling with robots with generation and recognition of facial expressions, \cite{bae2012towards} presents a promising starting point. Recognition of human facial expressions is accomplished through SHORE \cite{ruf2011face}, as well as the Seeing Machine’s product FaceAPI. Other available facial expression recognition systems include \cite{el2005real}, which has also been used as an aid for autistic children, as well as \cite{shan2009facial}, and \cite{bartlett2006fully}, where the output of the system is at the level of facial action coding (FACS). Regarding generation of facial expressions for robots, some examples of current research include \cite{wu2009learning}, \cite{han2013emotional}, \cite{ baltruvsaitis2010synthesizing} . Apart from static poses, the dynamics of facial expressions are also very important towards conveying believability; for empirical research on dynamics see for example \cite{littlewort2006dynamics}. Still, compared to the wealth of available research on the same subject with virtual avatars, there is still a lag both in empirical evaluations of human-robot affective interaction, as well as in importing existing tools from avatar animation towards their use for robots.

Regarding some basic supporting technologies of affect-enabled text-to-speech and speech recognition, the interested reader can refer to the general reviews by Schroeder \cite{schroder2009expressive} on TTS, and by Ververidis and Kotropoulos \cite{ververidis2006emotional} on recognition. A wealth of other papers on the subject exist; with some notable developments for affective speech-enabled real-world robotic systems including \cite{roehling2006towards} \cite{chella2008emotional}. Furthermore, if one moves beyond prosodic affect, to semantic content, the wide literature on sentiment analysis and shallow identification of affect applies directly; for example \cite{pang2008opinion} \cite{wilson2009recognizing} \cite{taboada2011lexicon}. Finally, regarding physiological measurables, products such as Affectiva’s Q sensor \cite{ picard2011measuring}, or techniques for measuring heart rate, breathing rate, galvanic skin response and more, could well become applicable to the human-robot affective interaction domain, of course under the caveats of \cite{ fairclough2009fundamentals}. Finally, it is worth noting that significant cross-culture variation exists regarding affect; both at the generation, as well as at the understanding and situational appropriateness levels \cite{ elfenbein2002universality}. In general, affective human-robot interaction is a growing field with promising results, which is expected to grow even more in the near future.

\subsection{Motor corellates of speech and non-verbal communication}\label{sec:motor}

Verbal communication in humans doesn’t come isolated from non-verbal signs; in order to achieve even the most basic degree of naturalness, any humanoid robot needs for example at least some lip-movement-like feature to accompany speech production. Apart from lip-syncing, many other human motor actions are intertwined with speech and natural language; for example, head nods, deictic gestures, gaze movements etc. Also, note that the term “corellates” is somewhat misleading; for example, the gesture channel can be more accurately described as being a complementary channel rather than a channel correlated with or just accompanying speech \cite{mcneill1992hand}. Furthermore, we are not interested only in the generation of such actions; but also on their combination, as well as on dialogic / interactional aspects.

Let us start by examining the generation of lip syncing. The first question that arises is: should lip sync actions be generated from phoneme-level information, or is the speech soundtrack adequate? Simpler techniques, rely on the speech soundtrack only; the simplest solution being to utilize only the loudness of the soundtrack, and map directly from loudness to mouth opening. There are many shortcomings in this approach; for example, a nasal ``m'' usually has large apparent loudness, although in humans it is being produced with a closed mouth.  Generally, the resulting lip movements of this method are perceivable unnatural. As an improvement to the above method, one can try to use spectrum matching of the soundtrack to a set of reference sounds, such as at \cite{weil1982face, lewis1984soft}, or even better, a linear prediction speech model, such as \cite{lewis1987automated}. Furthermore, apart from the generation of lip movements, their recognition can be quite useful regarding the improvement of speech recognition performance under low signal-to-noise ratio conditions \cite{bregler1994eigenlips}. There is also ample evidence that humans utilize lip information during recognition; a celebrated example is the McGurk effect \cite{mcgurk1976hearing}. The McGurk effect is an instance of so-called multi-sensory perception phenomena \cite{calvert2004handbook}, which also include other interesting cases such as the rubber hand illusion \cite{tsakiris2005rubber}.

Now, let us move on to gestures. The simplest form of gestures which are also directly relevant to natural language are deictic gestures, pointing towards an object and usually accompanied with indexicals such as ``this one!''. Such gestures have long been utilized in human-robot interaction; starting from virtual avatar systems such as Kris Thorisson’s Gandalf \cite{thorisson1996communicative} , and continuing all the way to robots such as ACE (Autonomous City Explorer) \cite{lidoris2009autonomous}, a robot that was able to navigate through Munich by asking pedestrians for directions. There exists quite a number of other types of gestures, depending on the taxonomy one adopts; such as iconic gestures, symbolic gestures etc. Furthermore, gestures are highly important towards teaching and learning in humans \cite{roth2001gestures}. Apart from McNeill’s seminal psychological work \cite{mcneill1992hand}, a definitive reference to gestures, communication, and their relation to language, albeit regarding virtual avatar Embodied Conversational Assistants (ECA), can be found in Justine Cassell’s work, including \cite{cassell1999embodiment, cassell2001beat}. Many open questions exist in this area; for example, regarding the synchronization between speech and the different non-verbal cues \cite{rossini2011patterns}, , and socio-pragmatic influences on the non-verbal repertoire. 

Another important topic for human-robot interaction is eye gaze coordination and hared attention. Eye gaze cues are important for coordinating collaborative tasks \cite{fussell2000coordination, brennan2008coordinating}, and also, eye gazes are an important subset of non-verbal communication cues that can increase efficiency and robustness in human-robot teamwork \cite{breazeal2005effects}. Furthermore, eye gaze is very important in disambiguating referring expressions, without the need for hand deixis \cite{hanna2007speakers, hanna2004pragmatic}. Shared attention mechanisms develop in humans during infancy \cite{adamson1991development}, and Scasellati authored the pioneering work on shared attention in robots in 1996 \cite{scassellati1996mechanisms}, followed up by \cite{scassellati1999imitation}. A developmental viewpoint is also taken in \cite{deak2001emergence}, as well as in \cite{fasel2002combining}. A well-cited probabilistic model of gaze imitation and shared attention is given in \cite{hoffman2006probabilistic}, In virtual avatars, considerable work has also taken place; such as \cite{peters2008towards, peters2010investigating}.

Eye-gaze observations are also very important towards mind reading and theory of mind \cite{premack1978does} for robots; i.e. being able to create models of the mental content and mental functions of other agents (human or robots) minds through observation. Children develop a progressively more complicated theory of mind during their childhood \cite{wellman2011child}. Elemental forms of theory of mind are very important also towards purposeful speech generation; for example, in creating referring expressions, one should ideally take into account the second-order beliefs of his conversational partner-listener; i.e. he should use his beliefs regarding what he thinks the other person believes, in order to create a referring expression that can be resolved uniquely by his listener. Furthermore, when a robot is purposefully issuing an inform statement (``there is a tomato behind you'') it should know that the human does not already know that; i.e. again an estimated model of second-order beliefs is required (i.e. what the robot believes the human believes). A pioneering work in theory of mind for robots is Scasellati’s \cite{scassellati2001foundations,scassellati2002theory}. An early implementation of perspective-shifting synthetic-camera-driven second-order belief estimation for the Ripley robot is given in \cite{mavridis2007grounded}. Another example of perspective shifting with geometric reasoning for the HRP-2 humanoid is given in \cite{marin2009towards}.

Finally, a quick note on a related field, which is recently growing. Children with Autistic Spectrum Disorders (ASD) face special communication challenges. A prominent theory regarding autism is hypothesizing theory-of-mind deficiencies for autistic individuals \cite{baron1997mindblindness, baron2000understanding}. However, recent research \cite{robins2005robotic, bird2007intact, robins2004robot, robins2009isolation} has indicated that specially-designed robots that interact with autistic children could potentially help them towards improving their communication skills, and potentially transferring over these skills to communicating not only with robots, but also with other humans.

Last but not least, regarding a wider overview of existing work on non-verbal communication between humans, which could readily provide ideas for future human-robot experiments, the interested reader is referred to \cite{vogeley2010artificial}.

\subsection{Purposeful speech and planning}\label{sec:purp_speech_planning}

Traditionally, simple command-only canned-response conversational robots had dialogue systems that could be construed as stimulus-response tables: a set of verbs or command utterances were the stimuli, the responses being motor actions, with a fixed mapping between stimuli and responses. Even much more advanced systems, that can support situated language, multiple speech acts, and perspective-shifting theory-of-mind, such as Ripley \cite{mavridis2007grounded}, can be construed as effectively being (stimulus, state) to response maps, where the state of the system includes the contents of the situation model of the robots. What is missing in all of these systems is an explicit modeling of purposeful behavior towards goals. 

Since the early days of AI, automated planning algorithms such as the classic STRIPS \cite{russell2009artificial} and purposeful action selection techniques have been a core research topic In traditional non-embodied dialogue systems practice, approaches such as Belief-Desire-Intention (BDI) have existed for a while \cite{jurafsky2000speech}, and theoretical models for purposeful generation of speech acts \cite{cohen1979elements} and computation models towards speech planning [BookSpeechPlanning] exist since more than two decades. Also, in robotics, specialized modified planning algorithms have mainly been applied towards motor action planning and path planning \cite{russell2009artificial}, such as RRT \cite{kuffner2000rrt} and Fast-Marching Squares \cite{garrido2006path}.

However, the important point to notice here is that, although considerable research exists for motor planning or dialogue planning alone, there are almost no systems and generic frameworks either for effectively combining the two, or for having mixed speech- and motor-act planning, or even better agent- and object-interaction-directed planners. Notice that motor planning and speech planning cannot be isolated from one another in real-world systems; both types of actions are often interchangeable with one another towards achieving goals, and thus should not be planned by separate subsystems which are independent of one another. For example, if a robot wants to lower its temperature, it could either say: “can you kindly open the window?” to a human partner (speech action), or could move its body, approach the window, and close it (motor action). An exemption to this research void of mixed speech-motor planning is \cite{mavridis2012ask}, where a basic purposeful action selection system for question generation or active sensing act generation is described, implemented on a real conversation robot. However, this is an early and quite task-specific system, and thus much more remains to be done towards real-world general mixed speech act and motor act action selection and planning for robots.

\subsection{Multi-level learning}\label{sec:learning}

Yet another challenge towards fluid verbal and non-verbal human-robot communication is concerned with learning \cite {klingspor1997human}. But when could learning take place, and what could be and should be learnt? Let us start by examining the when. Data-driven learning can happen at various stages of the lifetime of a system: it could either take place a) initially and offline, at design time; or, it could take place b) during special “learning” sessions, where specific aspects and parameters of the system are renewed; or, c) it could take place during normal operation of the system, in either a human-directed manner, or ideally d) through robot-initiated active learning during normal operation. Most current systems that exhibit learning, are actually involving offline learning, i.e. case a) from above. No systems in the literature have exhibited non-trivial online, real-world continuous learning of communications abilities.

The second aspect beyond the “when”, is the “what” of learning. What could be ideally, what could be practically, and what should be learnt, instead of pre-coded, when it comes to human-robot communication? For example, when it comes to natural-language communication, multiple layers exist: the phonological, the morphological, the syntactic, the semantic, the pragmatic, the dialogic. And if one adds the complexity of having to address the symbol grounding problem, a robot needs to have models of grounded meaning, too, in a certain target space, for example in a sensorymotor or a teleological target space. This was already discussed in the previous sections of “normative vs. empirical meaning” and on “symbol grounding”. Furthermore, such models might need to be adjustable on the fly; as discussed in the section on online negotiation of meaning. Also, many different aspects of non-verbal communication, from facial expressions to gestures to turn-taking, could ideally be learnable in real operation, even more so for the future case of robots needing to adapt to cultural and individual variations in non-verbal communications. Regarding motor aspects of such non-verbal cues, existing methods in imitation and demonstration learning \cite{argall2009survey} have been and could further be readily adapted; see for example the imitation learning of human facial expressions for the Leonardo robot \cite{breazeal1999imitation}.

Finally, another important caveat needs to be spelled out at this point. Real-world learning and real-world data collection towards communicative behavior learning for robots, depending on the data set size required, might require many hours of uninterrupted operation daily by numerous robots: a requirement which is quite unrealistic for today’s systems. Therefore, other avenues need to be sought towards acquiring such data sets; and crowdsourcing through specially designed online games offers a realistic potential solution, as mentioned in the previous paragraph on real-world acquisition of large-scale models of grounding. And of course, the learning content of such systems can move beyond grounded meaning models, to a wider range of the “what” that could be potentially learnable. A relevant example from a non-embodied setting comes from \cite{isbell2006cobot}, where a chatterbot acquired interaction capabilities through massive observation and interaction with humans in chat rooms. Of course, there do exist inherent limitations in such online systems, even for the case of the robot-tailored online games such as \cite{depalma2011leveraging}; for example, the non-physicality of the interaction presents specific obstacles and biases. Being able to extend this promising avenue towards wider massive data-driven models, and to demonstrate massive transfer of learning from the online systems to real-world physical robots, is thus an important research avenue for the future.

\subsection{Utilization of online resources and services}\label{sec:online}
 
Yet another interesting avenue towards enhanced human-robot communication that has opened up recently is the following: as more and more robots nowadays can be constantly connected to the internet, not all data and programs that the robot uses need to be onboard its hardware. Therefore, a robot could potentially utilize online information as well as online services, in order to enhance its communication abilities. Thus, the intelligence of the robot is partially offloaded to the internet; and potentially, thousands of programs and/or humans could be providing part of its intelligence, even in real-time. For example, going much beyond traditional cloud robotics \cite{guizzo2011robots}, in the human-robot cloud proposal \cite{mavridis2012human}, one could construct on-demand and on-the-fly distributed robots with human and machine sensing, actuation, and processing components. 

Beyond these highly promising glimpses of a possible future, there exist a number of implemented systems that utilize information and/or services from the internet. A prime example is Facebots, which are physical robots that utilize and publish information on Facebook towards enhancing long-term human-robot interaction, are described in \cite{mavridis2010facebots} \cite{mavridis2009facebots},. Facebots are creating shared memories and shared friends with both their physical as well as their online interaction partners, and are utilizing this information towards creating dialogues that enable the creation of a longer-lasting relationship between the robot and its human partners, thus reversing the quick withdrawal of the novelty effects of long-term HRI reported in \cite{mitsunaga2008makes}. Also, as reported in \cite{mavridis2011transforming}, the multilingual conversational robot Ibn Sina \cite{mavridis2009ibnsina}, has made use of online google translate services, as well as wikipedia information for its dialogues. Furthermore, one could readily utilize online high-quality speech recognition and text-to-speech services for human-robot communication, such as [Sonic Cloud online services], in order not to sacrifice onboard computational resources. 

Also, quite importantly, there exists the European project Roboearth \cite{ waibel2011roboearth}, which is described as “…a World Wide Web for robots: a giant network and database repository where robots can share information and learn from each other about their behavior and their environment. Bringing a new meaning to the phrase “experience is the best teacher”, the goal of RoboEarth is to allow robotic systems to benefit from the experience of other robots, paving the way for rapid advances in machine cognition and behaviour, and ultimately, for more subtle and sophisticated human-machine interaction”. Rapyuta \cite{hunziker2013rapyuta}, which is the cloud engine of Roboearth, claims to make immense computational power available to robots connected to it. Of course, beyond what has been utilized so far, there are many other possible sources of information and/or services on the internet to be exploited; and thus much more remains to be done in the near future in this direction.

\subsection{Miscellaneous abilities}\label{sec:misc}

Beyond the nine desiderata examined so far, there exist a number of other abilities that are required towards fluid and general human-robot communication. These have to do with dealing with multiple conversational partners in a discussion, with support for multilingual capabilities, and with generating and recognizing natural language across multiple modalities: for example not only acoustic, but also in written form. In more detail:

\subsubsection{Multiple conversational partners}

Regarding conversational turn-taking, in the words of Sacks \cite{sacks1974simplest}, “The organization of taking turns to talk is fundamental to conversation, as well as to other speech-exchange systems”, and this readily carries over to human-robot conversations, and becomes especially important in the case of dialogues with multiple conversation partners. Recognition of overlapping speech is also quite important towards turn-taking \cite{schegloff2000overlapping}. Regarding turn-taking in robots, a computational strategy for robots participating in group conversation is presented in \cite{matsusaka2001modeling}, and the very important role of gaze cues in turn taking and participant role assignment in human-robot conversations is examined in \cite{mutlu2009footing}. In \cite{chao2010turn}, an experimental study using the robot Simon is reported, which is aiming towards showing that the implementation of certain turn-taking cues can make interaction with a robot easier and more efficient for humans. Head movements are also very important in turn-taking; the role of which in keeping engagement in an interaction is explored in \cite{sidner2005explorations}.

Yet another requirement for fluid multi-partner conversations is sound-source localization and speaker identification. Sound source localization is usually accomplished using microphone arrays, such as the robotic system in \cite{valin2003robust}. An approach utilizing scattering theory for sound source localization in robots is described in \cite{nakadai2003applying} and approaches using beamforming for multiple moving sources are presented in \cite{valin2004localization} and \cite{valin2007robust}. Finally, HARK, an open-source robot audition system supporting three simultaneous speakers, is presented in \cite{nakadai2010design}. Speaker identification is an old problem; classic approaches utilize Gaussian mixture models, such as \cite{reynolds1995robust} and \cite{reynolds2000speaker}. Robotic systems able to identify their speaker’s identity include \cite{ji2008text}, \cite{krsmanovic2006have}, as well as the well-cited \cite{matsusaka1999multi}. Also, an important idea towards effective signal separation between multiple speaker sources in order to aid in recognition, is to utilize both visual as well as auditory information towards that goal. Classic examples of such approaches include \cite{nakadai2004improvement}, as well as \cite{katzenmaier2004identifying}. 

\subsubsection{Multilingual capabilities and Mutimodal natural language}

Yet another desirable ability for human-robot communication is multilinguality. Multilingual robots could not only communicate with a wider range of people, especially in multi-cultural societies and settings such as museums, but could very importantly also act as translators and mediators.  Although there has been considerable progress towards non-embodied multilingual dialogue systems \cite{holzapfel2005towards}, and multi-lingual virtual avatars do exist \cite{cullen2009reusable} \cite{echavarria2005multilingual}, the only implemented real-world multilingual physical android robot so far reported in the literature is \cite{mavridis2011transforming}. 

Finally, let us move on to examining multiple modalities for the generation and recognition of natural language. Apart from a wealth of existing research on automated production and recognition of sign language for the deaf (ASL) \cite{ starner1998real} \cite{vogler2004handshapes} \cite{murthy2009review}, systems directly adaptable to robots also exist \cite{brashear2003using}. One could also investigate the intersection between human writing and robotics. Again, a wealth of approaches exist for the problem of optical character recognition and handwriting recognition \cite{plamondon2000online} \cite{plotz2009markov}, even for languages such as Arabic \cite{lorigo2006offline}, the only robotic system that has demonstrated limited OCR capabilities is \cite{mavridis2011transforming}. Last but not least, another modality available for natural language communication for robots is internet chat. The only reported system so far that could perform dialogues both physically as well as through facebook chat is \cite{mavridis2010facebots} \cite{mavridis2009facebots}. 

As a big part of human knowledge, information, as well as real-world communication is taking place either through writing or through such electronic channels, inevitably more and more systems in the future will have corresponding abilities. Thus, robots will be able to more fluidly integrate within human societies and environments, and ideally will be enabled to utilize the services offered within such networks for humans. Most importantly, robots might also one day become able to help maintain and improve the physical human-robot social networks they reside within towards the benefit of the common good of all, as is advocated in \cite{mavridis2011autonomy}.

\section{Discussion}\label{sec:discussion}

From our detailed examination of the ten desiderata, what follows first is that although we have moved beyond the “canned-commands-only, canned responses” state-of-affairs of the ninetees, we seem to be still far from our goal of fluid and natural verbal and non-verbal communication between humans and robots. But what is missing? 

Many promising future directions were mentioned in the preceeding sections. Apart from clearly open avenues for projects in a number of areas, such as composition of grounded semantics, online negotiation of meaning, affective interaction and closed-loop affective dialogue, mixed speech-motor planning, massive acquisition of data-driven models for human-robot communication through crowd-sourced online games, real-time exploitation of online information and services for enhanced human-robot communication, many more open areas exist.

What we speculate might really make a difference, though, is the availability of massive real-world data, in order to drive further data-driven models. And in order to reach that state, a number of robots need to start getting deployed, even if in partially autonomous partially remote-human-operated mode, in real-world interactive application settings with round-the-clock operation: be it shopping mall assistants, receptionists, museum robots, or companions, the application domains that will bring out human-robot communication to the world in more massive proportions, remains yet to be discovered. However, given recent developments, it does not seem to be so far away anymore; and thus, in the coming decades, the days might well come when interactive robots will start being part of our everyday lives, in seemless harmonious symbiosis, hopefully helping create a better and exciting future.

\section{Conclusions}\label{sec:conclusions}

An overview of research in human-robot interactive communication was presented, covering verbal as well as non-verbal aspects. Following a historical introduction reaching from roots in antiquity to well into the ninetees, and motivation towards fluid human-robot communication, ten desiderata were proposed, which provided an organizational axis both of recent as well as of future research on human-robot communication. Then, the ten desiderata were explained, relevant research was examined in detail, culminating to a unifying discussion. In conclusion, although almost twenty-five years in human-robot interactive communication exist, and significant progress has been achieved in many fronts, many sub-problems towards fluid verbal and non-verbal human-robot communication remain yet unsolved, and present highly promising and exciting avenues towards research in the near future.

\bibliographystyle{./IEEEtran} 
\bibliography{robots}

\end{document}